\documentclass[sigconf]{acmart}

\usepackage{latexsym}
\usepackage[utf8]{inputenc}
\usepackage{microtype}

\PassOptionsToPackage{table}{xcolor}
\usepackage{colortbl}
\usepackage{subfigure}

\DeclareMathOperator*{\argmax}{arg\,max}

\newcommand{\dorange}[1]{\textcolor{orange}{#1}}
\newcommand{\doblue}[1]{\textcolor{blue}{#1}}

\usepackage{bibentry}
\usepackage{arydshln}
\usepackage{tabularray}
\definecolor{mgreen}{rgb}{0.0, 0.5, 0.0}
\newcommand{\dogreen}[1]{\textcolor{mgreen}{#1}}

\usepackage[most]{tcolorbox}

\definecolor{headercolor}{gray}{0.85}
\definecolor{boxcolor}{gray}{0.95}

\definecolor{lightgray}{gray}{0.95}
\definecolor{darkgray}{gray}{0.85}
\definecolor{highlight}{gray}{0.9}
\definecolor{questioncolor}{gray}{0.9} 
\definecolor{answercolor}{gray}{0.95}   

\colorlet{Mycolor1}{red!30}
\usepackage{etoolbox}
\AtBeginEnvironment{tcolorbox}{\small}

\AtBeginDocument{%
  }

\copyrightyear{2024} 
\acmYear{2024} 
\setcopyright{acmlicensed}\acmConference[ICAIF '24]{5th ACM International Conference on AI in Finance}{November 14--17, 2024}{Brooklyn, NY, USA}
\acmBooktitle{5th ACM International Conference on AI in Finance (ICAIF '24), November 14--17, 2024, Brooklyn, NY, USA}
\acmDOI{10.1145/3677052.3698690}
\acmISBN{979-8-4007-1081-0/24/11}

\acmConference[ICAIF'24]{5th ACM International Conference on AI in Finance (ICAIF-24)}{14-16 November,
  2024}{Brooklyn, NY}
\acmISBN{978-1-4503-XXXX-X/18/06}

\begin{document}

\title{TADACap: Time-series Adaptive Domain-Aware Captioning}

\author{Elizabeth Fons}
\email{elizabeth.fons@jpmorgan.com}
\affiliation{
    \institution{JP Morgan AI Research}
    \city{London}
    \country{UK}
}

\author{Rachneet Kaur}
\email{rachneet.kaur@jpmorgan.com}
\affiliation{
    \institution{JP Morgan AI Research}
    \city{New York}
    \state{New York}
    \country{USA}
}

\author{Zhen Zeng}
\email{zhen.zeng@jpmorgan.com}
\affiliation{
    \institution{JP Morgan AI Research}
    \city{New York}
    \state{New York}
    \country{USA}
}

\author{Soham Palande}
\email{soham.palande@jpmorgan.com}
\affiliation{
    \institution{JP Morgan AI Research}
    \city{New York}
    \state{New York}
    \country{USA}
}

\author{Tucker Balch}
\email{tucker.balch@jpmorgan.com}
\affiliation{
    \institution{JP Morgan AI Research}
    \city{New York}
    \state{New York}
    \country{USA}
}

\author{Svitlana Vyetrenko}
\email{svitlana.vyetrenko@jpmorgan.com}
\affiliation{
    \institution{JP Morgan AI Research}
    \city{San Francisco}
    \state{California}
    \country{USA}
}

\author{Manuela Veloso}
\email{manuela.veloso@jpmorgan.com}
\affiliation{
    \institution{JP Morgan AI Research}
    \city{New York}
    \state{New York}
    \country{USA}
}

\renewcommand{\shortauthors}{Fons et al.}

\begin{abstract}
While image captioning has gained significant attention, the potential of captioning time-series images, prevalent in areas like finance and healthcare, remains largely untapped. Existing time-series captioning methods typically offer generic, domain-agnostic descriptions of time-series shapes and struggle to adapt to new domains without substantial retraining. To address these limitations, we introduce TADACap, a retrieval-based framework to generate domain-aware captions for time-series images, capable of adapting to new domains without retraining. Building on TADACap, we propose a novel retrieval strategy that retrieves diverse image-caption pairs from a target domain database, namely TADACap-diverse. We benchmarked TADACap-diverse against state-of-the-art methods and ablation variants. TADACap-diverse demonstrates comparable semantic accuracy while requiring significantly less annotation effort. 
\end{abstract}



\maketitle

\section{Introduction}

Image captioning \cite{radford2021learning, wang2021simvlm, hu2022scaling, li2022blip, wang2022git} has gained significant attention and advances in the field of computer vision and natural language processing. The use of image captioning spans widely from aiding visually impaired individuals, and automating content discovery, to enhancing user experience in digital media. The majority of image captioning works focus on images of scenery or objects, leaving a vast area of potential unexplored, specifically in the captioning of time-series images.

Time-series plots are commonly used in many domains, such as finance, healthcare, climate science, and business analytics, to display trends and patterns in data over time. Despite their prevalence, automatic captioning of time-series is under-explored compared to captioning of natural images. Recent advances have been made in time-series captioning, some works focus on captioning raw time-series data \cite{andreas2014grounding, sowdaboina2014learning, murakami2017learning, jhamtani2021truth}, and others explore captioning time-series images \cite{chen2019figure, qian2021generating, mahinpei2022linecap}. We argue that \textbf{image-based} time-series captioning has a broader application range because it does not require access to the raw time-series data, such as when captioning time-series plots in reports. 

\begin{figure}[t!]
    \centering
    \includegraphics[width=0.47\textwidth]{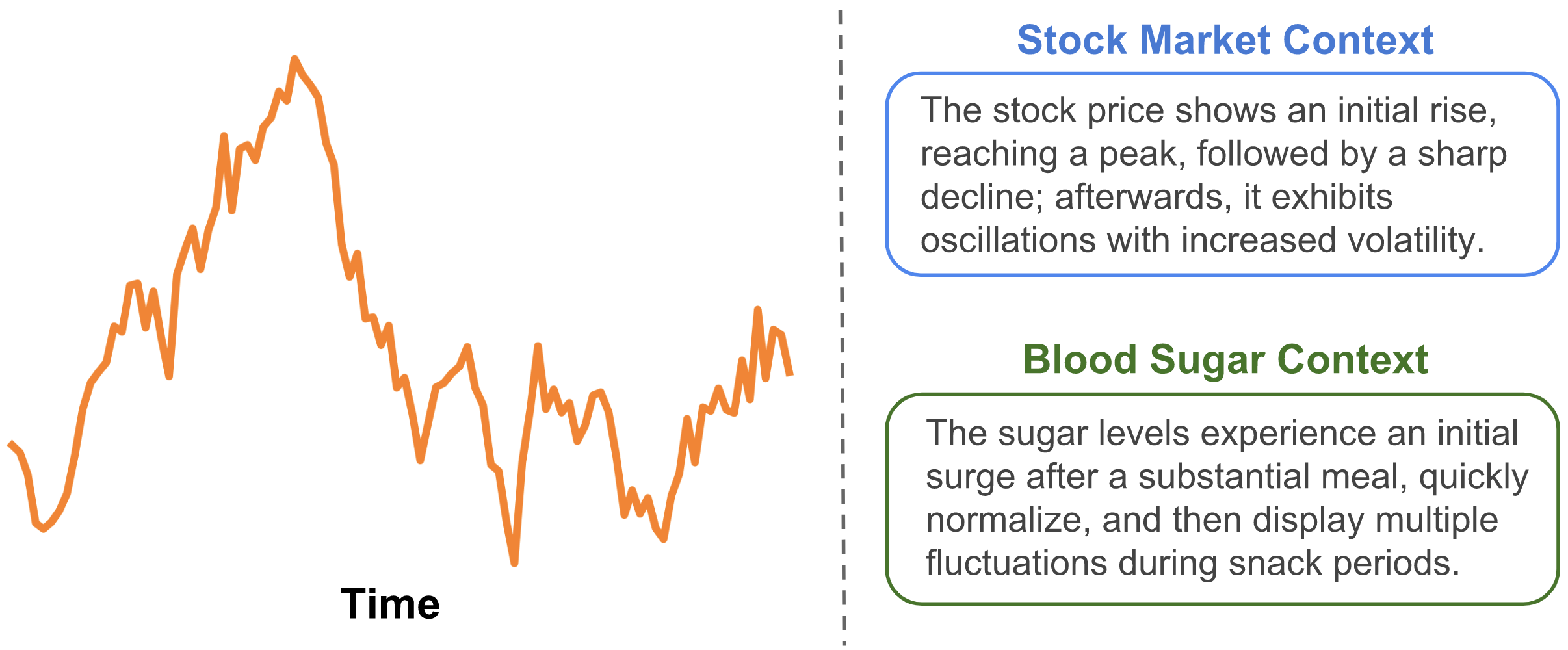}
    \caption{Motivation of domain-aware time-series captioning. The caption of a given time-series in one domain can be drastically different from the caption of the same (or similarly shaped) time-series in another domain.}
    \label{fig:teaser}
\end{figure}

Existing methods for time-series captioning, regardless of whether raw time-series based or image-based, typically generate generic, domain-agnostic captions of time-series shapes. These captions lack the contextual relevance and domain specificity crucial for meaningful interpretation. As illustrated in Figure~\ref{fig:teaser}, the corresponding captions differ greatly when a trader in the finance domain observes the same time-series versus a doctor in the healthcare domain. The same or similarly shaped time-series have different implications and meanings in different domains.
In this paper, we tackle this challenge of domain-aware captioning for time-series.

There exist recent works on image captioning adapted for specific domains \cite{hu2022scaling, li2022blip, wang2022git, li2020oscar, wang2021simvlm}. However, these works require training or finetuning to adapt to specific domains, and struggle to adapt to new domains without extensive retraining. This poses a significant limitation in their practical applicability.

To address these challenges, we introduce TADACap -- a retrieval-based framework capable of generating domain-aware captions for time-series images and adapting to new domains without retraining. In this framework, we propose a novel and effective retrieval strategy that retrieves diverse samples from the target domain, in contrast to common practices of nearest-neighbor retrieval, named as TADACap-diverse. To the best of our knowledge, \textbf{our work is the first} to effectively tackle these challenges, setting it apart from existing methods. TADACap is simple but effective, it runs with off-the-shelf models without the requirement for finetuning. TADACap-diverse leverages diverse image-caption pairs from a target domain database to produce contextually relevant captions for query images. 
Our contributions are three-fold:
\begin{itemize}
    \item We propose TADACap, a retrieval-based captioning framework that adapts to new domains and generates domain-aware captions for time-series images without retraining on the new domains.
    \item We propose TADACap-diverse, i.e., TADACap with the proposed diverse retrieval strategy, only requires caption annotations of a few diverse image samples for domain-aware captioning, while maintaining comparable semantic accuracy. Thereby largely reducing required annotation effort compared to other retrieval-based methods \cite{sarto2022retrieval, ramos2023retrieval, ramos2022smallcap}.
    \item We introduce new datasets for domain-aware time-series captioning.
\end{itemize}

The rest of the paper is organized as follows: Section \ref{sec:related_works} discusses related models on image captioning, retrieval-based image captioning, and time-series captioning. Section \ref{methods} describes the methods and implementation details. Section \ref{experiments} covers datasets, benchmarks, and results. Finally, Section \ref{conclusions} provides the conclusions.
\section{Related work}\label{sec:related_works}

\subsection{Image Captioning}
Image captioning approaches commonly adopted involve the use of encoder-decoder methods \cite{anderson2018bottom, xu2015show, yang2022reformer}. In these methods, an input image is passed through a visual encoder, and a caption is generated using an autoregressive language decoder. Some recent benchmarks in image captioning are defined by large-scale general-purpose vision and language (V\&L) models \cite{hu2022scaling, li2022blip, wang2022git, li2020oscar, wang2021simvlm}, which are pre-trained on a large number of image-text pairs to learn generic features and then fine-tuned for specific downstream tasks such as image captioning.
Despite the impressive performance of these V\&L models on natural images from different domains, adapting them for image captioning in new domains can be time-consuming and expensive, as it often requires a separate model to be fine-tuned or optimized for each new dataset \cite{agrawal2019nocaps, gurari2020captioning}. Further, as these models are scaled up, the computational requirements for pre-training and fine-tuning on downstream tasks also increase.

Recent works, such as ClipCap \cite{mokady2021clipcap}, I-Tuning \cite{luo2022tuning} and SmallCap \cite{ramos2022smallcap}, have addressed these issues of expensive and time-consuming fine-tuning on new data by utilizing pre-trained vision encoder, CLIP \cite{radford2021learning}, and language decoder, GPT-2 \cite{radford2019language}, as frozen model components. ClipCap uses prefix-tuning to map a fixed-length CLIP embedding of the image into the GPT-2 language space. I-Tuning adjusts the output hidden states of GPT-2 by extracting visual memory embeddings from CLIP. SmallCap uses trainable cross-attention layers to connect CLIP and GPT-2, and retrieval augmentation (Section~\ref{sec:retrieval_cap}) to reduce the number of trainable parameters while maintaining performance. While these models have shown promising results on natural images, the use of images for time-series captioning is under-explored. 
More recently, multimodal LLMs have become state-of-the-art for most tasks, including image captioning \cite{liu2024visual, ye2023mplugowl, achiam2023gpt, team2023gemini}.

\subsection{Retrieval-based Image Captioning} \label{sec:retrieval_cap}
To avoid extensive training, retrieval-based image captioning has been studied~\cite{ramos2021retrieval, xu2019unified, zhao2020image}, which involves conditioning the generation of captions on additional information retrieved from an external datastore \cite{lewis2020retrieval}. More recent works \cite{sarto2022retrieval, ramos2023retrieval} have introduced transformer-based captioning models that use retrieved captions as additional information for generating captions, and perform cross-attention over the encoded retrieved captions. Although the previously mentioned works might have the potential to adapt to new domains, they have not been tested on domain adaptation. On the other hand, SmallCap~\cite{ramos2022smallcap} introduces a novel prompt-based conditioning approach, wherein retrieved captions are utilized as a prompt for a generative language model. It should be noted that SmallCap is the first work that leverages retrieval augmentation for training-free domain transfer and generalization in image captioning. Further, some recent advancements in retrieval-augmented generation are highlighted by \cite{yasunaga2022retrieval, zhao2024retrieval, ding2024survey}, which explore the integration of retrieval mechanisms with language models to enhance performance and accuracy across various tasks.

Our proposed method in image captioning utilizes prompts with a task demonstration, much like SmallCap, which is currently the state-of-the-art in retrieval-based image captioning with domain adaptation. However, SmallCap requires a fully annotated target domain database to retrieve relevant information, which is not a requirement for our method. This will be demonstrated and discussed in our experiments.

\begin{figure*}[t!]
    \centering
    \includegraphics[clip, trim=0cm 0cm 0cm 0.7cm,width=0.95\textwidth]{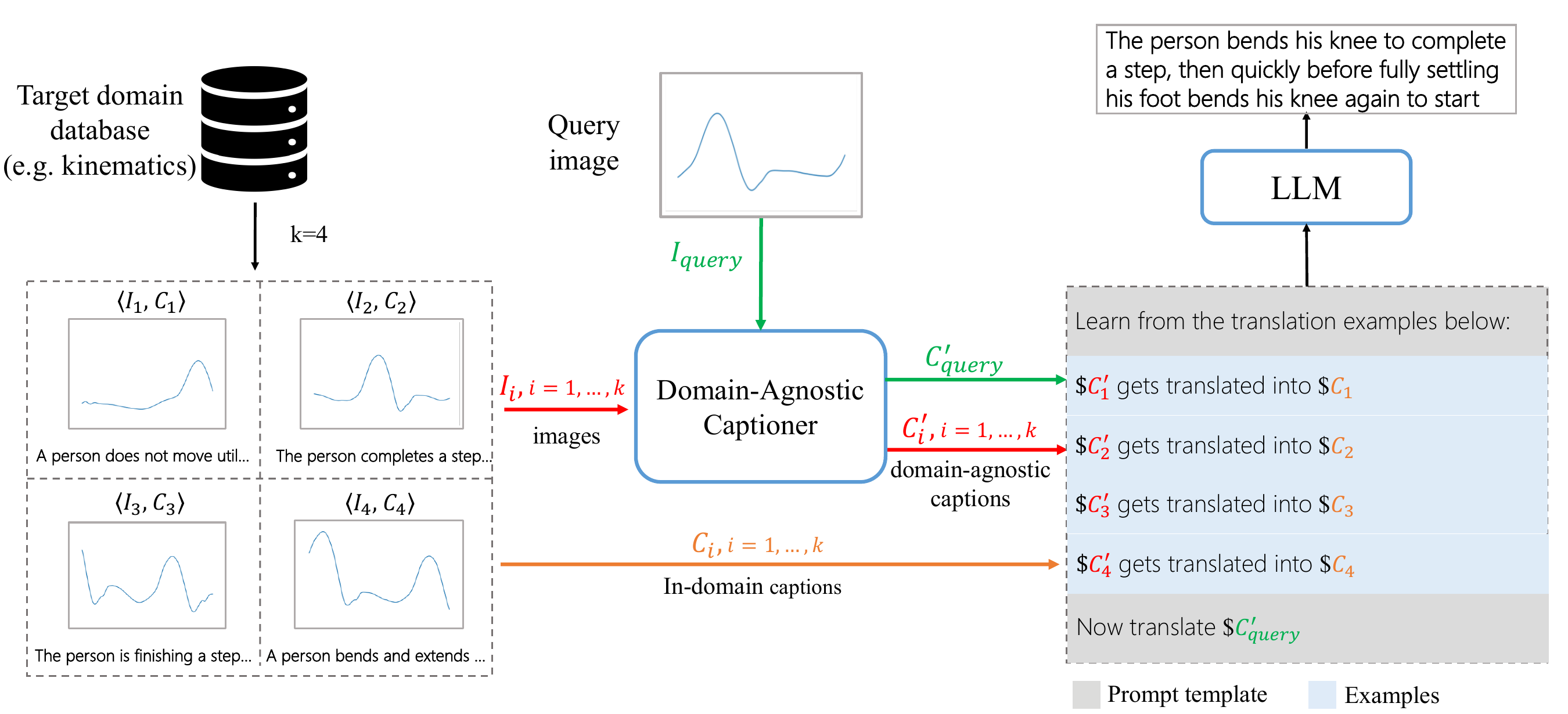}
    \caption{Overview of proposed TADACap framework for domain-adaptive time-series captioning. TADACap generates an in-domain caption of a query image based on a retrieved set of $k$ domain-agnostic and in-domain caption pairs, which are used as examples in the prompt to the GPT decoder. When adopting the proposed strategy of retrieving diverse samples from the target domain database, we name our full proposed approach as TADACap-diverse.}
    \label{fig:overview}
\end{figure*}

\subsection{Time-series Captioning} \label{sec:tscap}
One approach to time-series captioning is to use raw numerical time-series data, as explored in several works including \cite{andreas2014grounding, sowdaboina2014learning, murakami2017learning, jhamtani2021truth}. However, in some cases, only plots of the time-series data are available, such as in reports or presentations \cite{kahou2017figureqa}. This presents a challenge for downstream time-series tasks, such as generating captions, as the raw data is not accessible. To address this challenge, several recent works have used images for time-series forecasting \cite{sood2021visual, zeng2021deep, semenoglou2023image, zeng2023pixels} and classifications tasks \cite{li2023time, cohen2020trading}. In the context of time-series captioning, researchers have specifically focused on automatically generating captions for time-series plots, as explored in studies including \cite{chen2019figure, qian2021generating, mahinpei2022linecap}. Similar to LineCap \cite{mahinpei2022linecap}, we aim to tackle time-series image captioning for a broad range of application scenarios. However, all existing work in time-series captioning, whether image-based or raw time-series-based, generates captions that describe the generic shape of the time-series. They do not take into account the domain-aware features of the time-series; thus, captions from these models suffer from a lack of contextual relevance and domain specificity, which is essential for interpreting them meaningfully.

In this paper, we propose a novel approach to time-series captioning that is domain-aware. We are the first to tackle this problem, and our approach promises to generate captions that are beyond generic shape descriptions and specific to the domain.
\section{Method}\label{methods}

We propose the framework Time-series Adaptive Domain-Aware Captioning (TADACap) to tackle the problem of domain-aware captioning of time-series, without retraining, as shown in Figure~\ref{fig:overview}. We refer to our generic retrieval-based captioning framework as TADACap. Building on TADACap, when adopting the proposed diverse retrieval strategy, we refer to it as \textbf{TADACap-diverse} as the full name of our proposed approach.

Given a query time-series image $I_{query}$, TADACap first uses a \textit{Domain-Agnostic Captioner} to generate a domain-agnostic caption $C_{query}$ that describes shape of the query time-series. From the target domain database, TADACap with our proposed diverse retrieval strategy, namely TADACap-diverse, retrieves a diverse set of $k$ time-series images and in-domain caption pairs $\langle I_i, C_i \rangle$, as explained in detail in Section~\ref{sec:diverse_k}. Then TADACap generates a domain-agnostic caption $C_i^\prime$ corresponding to each of those retrieved $k$ time-series images, thus composing a list of in-domain and domain-agnostic caption pairs $\langle C_i, C_i^\prime \rangle$. 
Finally, TADACap adapts \(C_{\text{query}}\) to the target domain through prompts to a Large Language Model (LLM) containing \(\langle C_i, C_i' \rangle\), as in in-context learning (ICL). The prompt template is as shown in Figure~\ref{fig:overview}.

The \textit{Domain-Agnostic Captioner} generates a generic shape description of a given time-series (or time-series image). Note that our framework is generic in that we don't impose any particular architecture constraints on the \textit{Domain-Agnostic Captioner}, as long as it generates a shape description of a given time-series (or time-series image). This \textit{Domain-Agnostic Captioner} can be any image captioning methods~\cite{ramos2022smallcap, mokady2021clipcap, barraco2022unreasonable, tewel2021zero, luo2022tuning} pre-trained on time-series images with domain-agnostic captions. 

\subsection{Diverse $k$ Samples}\label{sec:diverse_k}
Given a target domain database, we aim to retrieve $k$ diverse samples from the database. 
One key innovation of our method is retrieving \textbf{diverse} samples from the target domain, which provides a good coverage of the target domain knowledge.

Existing works on retrieval-based captioning~\cite{ramos2022smallcap, ramos2021retrieval, xu2019unified, zhao2020image} assume that a fully annotated target domain database is available such that nearest-neighbor samples to the query image can be retrieved and leveraged. This assumption is valid in the domain of images of objects thanks to large public image datasets such as COCO~\cite{chen2015microsoft}. However, in the domain of times series, no such dataset with in-domain captions exists, thus we introduced such databases as discussed in Section~\ref{sec:datasets}.

Annotating captions of time-series can be expensive, especially for time-series that require domain knowledge such as sensor readings (e.g. Electrocardiogram (ECG) signals, satellite readings). One advantage of our method is \textbf{largely reduced annotation efforts} on image captioning for users to prepare the target domain database. Users can effectively construct a target domain database as follows, 1) first collect a set of time-series images from the target domain without captions; 2) use our method to select $k$ diverse samples from the database; 3) annotate the captions only for the selected $k$ samples rather than all samples in the database.

To achieve this, we first compute the CLIP image embedding of each time-series image in the target domain database; then, we use Determinantal Point Process (DPP)~\cite{kulesza2012determinantal} to select a diverse subset of $k$ samples from the set of embeddings. DPPs are probabilistic models that capture diversity by assigning a higher probability to subsets of points that are well spread out. DPPs have proven to be effective in discovering a diverse subset of samples given a collection of high-dimensional points.

Formally, we compute the embeddings of the time-series images from the target domain database as $X=\{x_1, x_2, \cdots, x_N\}$, and we define a kernel matrix $L \in \mathbb{R}^{N \times N}$ where $L(i,j) = sim(x_i, x_j)$ computes the cosine similarity between embedding $x_i$ and $x_j$. A DPP measures the probability of selecting a subset $S$ from $X$ as
\begin{equation}
    P(S) = \det(L(S)) / \det(L + I)
\end{equation}
where $\det(\cdot)$ computes the determinant of an input matrix, $I$ is the $N \times N$ identity matrix, where $N$ is the number of images in the database. To get the subset of size $k$ with the maximum diversity, we solve a maximum a posteriori (MAP) problem,
\begin{equation}
    S^* = \argmax_{S, |S|=k} \det(L(S)) / \det(L + I)
\end{equation}
which can be solved through an efficient greedy algorithm~\cite{chen2018fast} with complexity $O(Nk^2)$. It is not guaranteed to find the global maximum of the DPP distribution. However, in practice, it often produces high-quality subsets that are diverse and representative of the original set~\cite{celis2018fair, wilhelm2018practical, wang2020personalized, liu2021diversity}.

\begin{equation}
L_{\theta} = - \sum^N_{I=1} \log P_{\theta}(y_i|y_{<I}, \mathbf{X}, \mathbf{V}, \theta)    
\end{equation}

We used $k=4$ across our experiments. We determined $k=4$ based on iterating the Determinantal Point Process (DDP) diverse sample selection until adding a new sample does not significantly change the probability of the subset~\cite{chen2018fast}.
\subsection{Implementation}\label{sec:impl}

Note that TADACap is compatible with any image-based \textit{Domain-Agnostic Captioner} capable of generating domain-agnostic time-series captions. Training such captioner is not within the scope of the paper, thus we used an off-the-shelf domain-agnostic captioner in our experiments. Specifically, we prompted GPT-4V to describe the generic shape of the test time-series, without domain context.

\section{Experiments}\label{experiments}

We benchmarked the domain-aware time-series image captioning of the proposed TADACap with retrieval strategy on diverse samples, TADACap-diverse in particular, against TADACap ablation variants, as well as a state-of-the-art free multi-modal LLMs including Gemini-V~\cite{team2023gemini} and mPLUG-Owl~\cite{ye2023mplugowl}, as detailed in Section~\ref{sec:benchmarks}. We are the first to tackle domain-aware time series captioning, previous works that only generate domain-agnostic captions~\cite{chen2019figure, qian2021generating, mahinpei2022linecap, mahinpei2022linecap} are not benchmarked against due to their limited scope. We introduced four datasets for domain-aware time series captioning, ranging from finance, physics, and clinical, to kinematics. We demonstrated the overall success of our method across various domains.

\begin{figure*}[htb]
\centering
\includegraphics[width=0.93\textwidth]{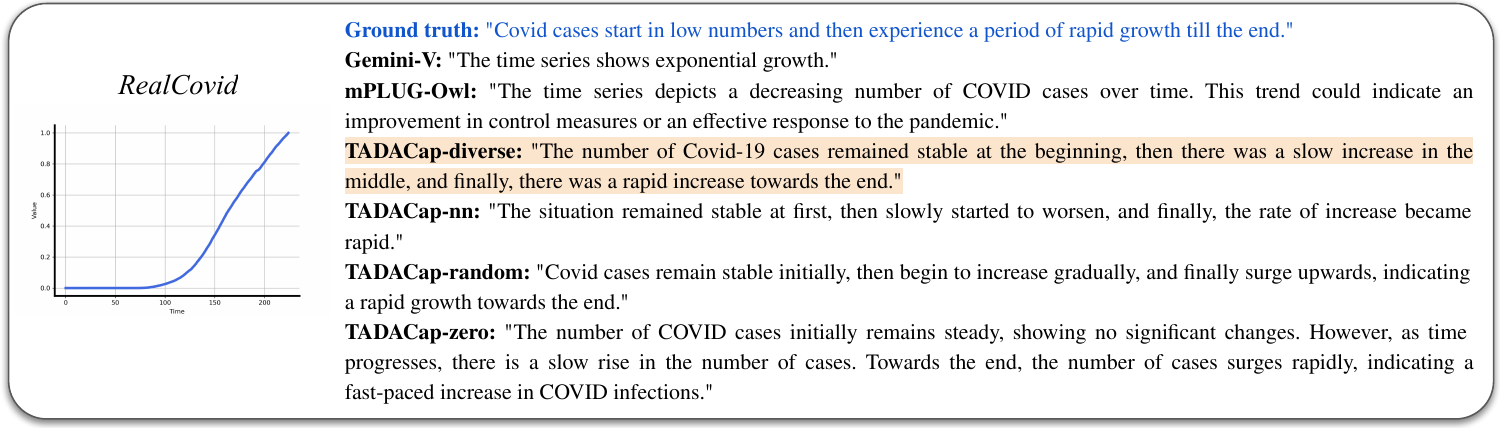}\\
\includegraphics[width=0.93\textwidth]{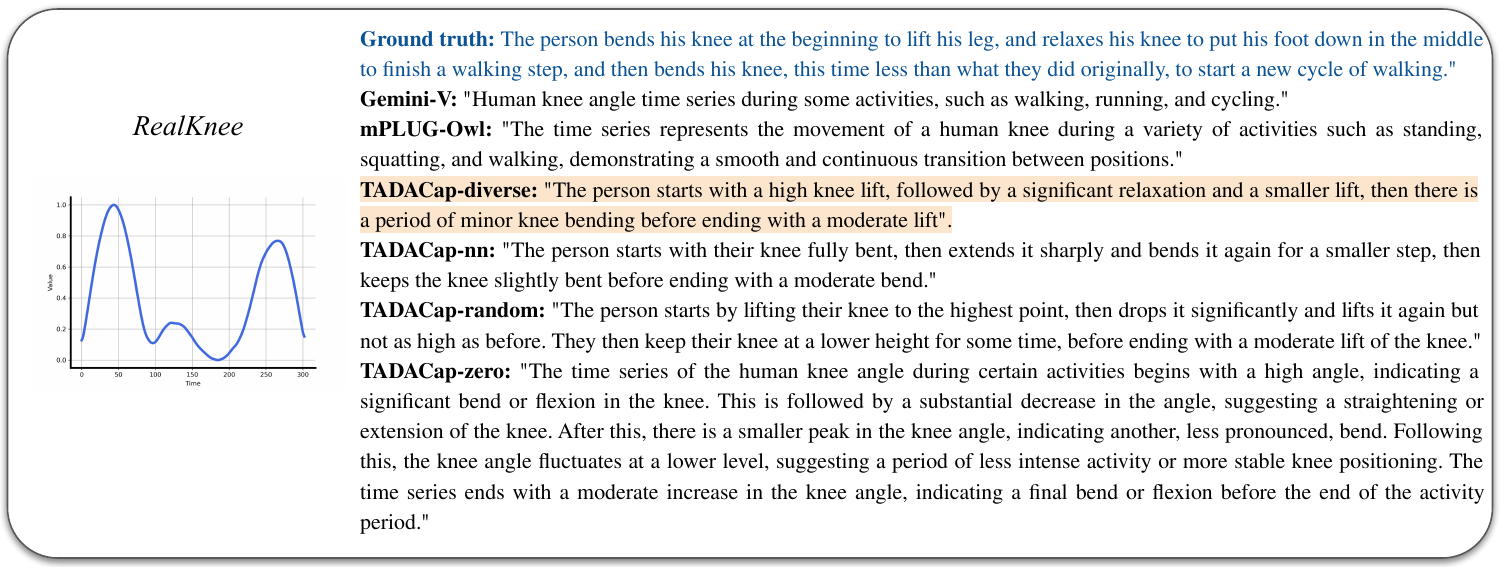}\\
\includegraphics[width=0.93\textwidth]{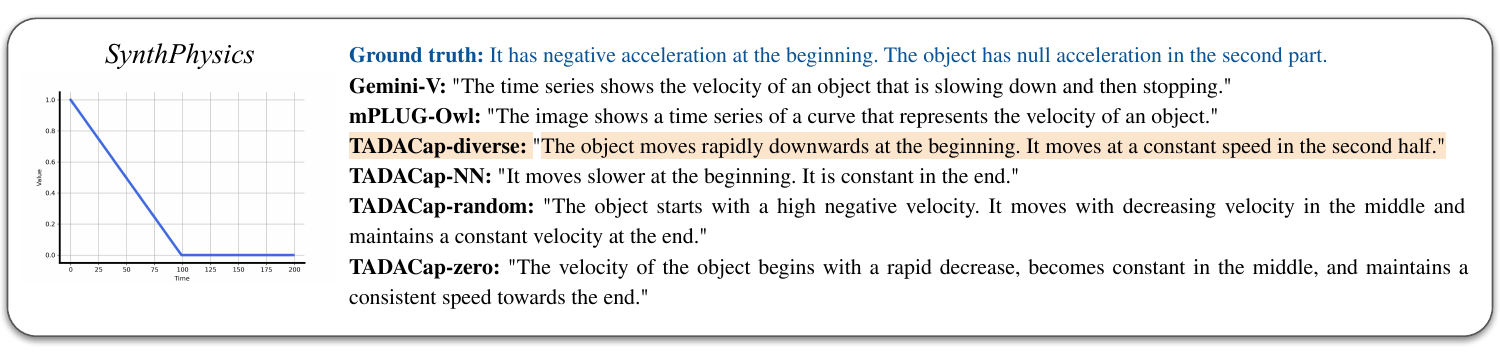}\\
\includegraphics[width=0.93\textwidth]{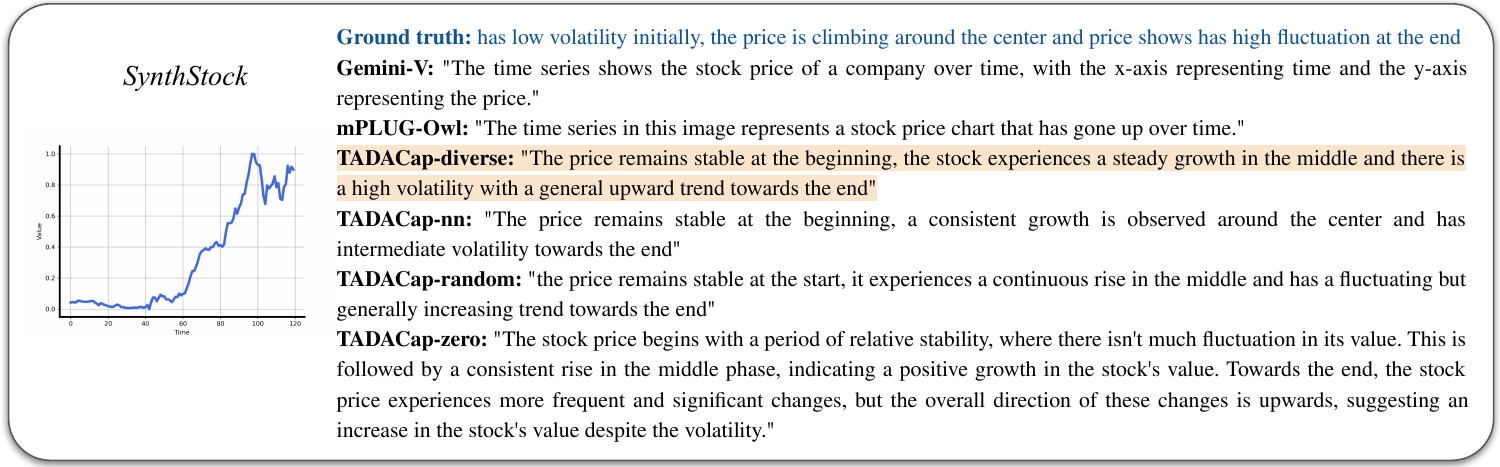}
\caption{Qualitative results on the \textit{RealCovid}, \textit{RealKnee}, \textit{SynthPhysics} and \textit{SynthStock} time-series dataset for domain-aware captioning.}
\label{fig:datasets_results}
\end{figure*}

\newcommand{\ctt}[2]{\colorbox{#1}{{#2}}}
\colorlet{pale1}{blue!10}
\colorlet{pale2}{green!10}
\colorlet{pale3}{red!10}
\colorlet{pale4}{orange!10}
\colorlet{pale5}{cyan!10}
\colorlet{pale6}{magenta!10}
\colorlet{pale7}{gray!10}
\colorlet{pale8}{teal!10}
\colorlet{pale9}{purple!10}

\cellcolor{pale7}

\begin{table*}[]
    \centering
    \parbox{0.49\linewidth}{
        \resizebox{\linewidth}{!}{
            \begin{tabular}{llcccc}
                \toprule
                & & ROUGE-L & CIDEr & SPICE & SPIDEr \\
                \midrule
                \textit{RealCovid} & & & & \\
                \midrule
                Multimodal LLM & Gemini-V & 20.3 & 7.3 & 19.5 & 13.4 \\
                Multimodal LLM & mPLUG-Owl & 22.3 & 3.9 & 15.4 & 9.7 \\
                \midrule
                \textbf{TADACap-diverse} & GPT4 & \cellcolor{pale1} \bf 26.5 & \cellcolor{pale7} 8.1 & \cellcolor{pale1} \bf 26.2 & \cellcolor{pale1} \bf 17.1 \\
                TADACap-nn & GPT4 & \cellcolor{pale7} 25.9 & \cellcolor{pale1} \bf 9.8 & 22.7 & \cellcolor{pale7} 16.2 \\
                TADACap-random & GPT4 & 25.4 & 7.1 & 23.4 & 15.3 \\
                TADACap-zs & GPT4 & \cellcolor{pale7} 25.9 & 4.7 & \cellcolor{pale7} 23.5 & 14.1 \\
                \midrule
                \textbf{TADACap-diverse} & Vicuna & \cellcolor{pale1} 24.8 & \cellcolor{pale1} 8.3 & \cellcolor{pale1} 18.7 & \cellcolor{pale1} 13.5 \\
                TADACap-nn  & Vicuna & \cellcolor{pale7} 24.6 & \cellcolor{pale7} 7.5 & \cellcolor{pale7} 17.2 & \cellcolor{pale7} 12.3 \\
                TADACap-random & Vicuna & 23.2 & 5.5 & 16.4 & 11.0 \\
                TADACap-zs & Vicuna & 19.9 & 4.4 & 14.5 & 9.5 \\
                \midrule
                \textit{SynthPhysics} & & & \\
                \midrule
                Multimodal LLM & Gemini-V & 19.0 & 11.3 & 16.6 & 14.0 \\
                Multimodal LLM & mPLUG-Owl & 20.1 & 5.1 & 12.8 & 9.0 \\
                \midrule
                \textbf{TADACap-diverse} & GPT4 & \cellcolor{pale7} 33.0 & \cellcolor{pale7} 30.3 & \bf \cellcolor{pale1} 20.3 &  \cellcolor{pale7} 25.3 \\
                TADACap-nn  & GPT4 & \cellcolor{pale1} \bf 33.5 & \cellcolor{pale1} 39.4 & \bf \cellcolor{pale1} 20.3 & \bf \cellcolor{pale1} 29.9 \\
                TADACap-random & GPT4 & 30.8 & 26.8 & 19.8 & 23.3 \\
                TADACap-zs & GPT4 & 23.7 & 7.4 & 18.0 & 12.7 \\
                \midrule
                \textbf{TADACap-diverse} & Vicuna & 27.2 & \cellcolor{pale7} 11.2 & \cellcolor{pale7} 16.1 & \cellcolor{pale7} 13.7 \\
                TADACap-nn  & Vicuna & \cellcolor{pale7} 27.8 & \cellcolor{pale1} 24.7 & 13.6 & \cellcolor{pale1} 19.2 \\
                TADACap-random & Vicuna & \cellcolor{pale1}  30.2 & 10.5 & \cellcolor{pale1} 18.8 & 14.6 \\
                TADACap-zs & Vicuna & 24.5 & 6.3 & 14.6 & 10.4 \\
                \bottomrule
            \end{tabular}
        }
    }
    \parbox{0.49\linewidth}{
        \resizebox{\linewidth}{!}{
            \begin{tabular}{llcccc}
            \toprule
            &  & ROUGE-L & CIDEr & SPICE & SPIDEr \\
            \midrule
            \textit{RealKnee} & & & & \\
            \midrule
            Multimodal LLM & Gemini-V & 14.7 & 1.7 & 5.5 & 3.6 \\
            Multimodal LLM & mPLUG-Owl & 15.5 & 1.3 & 5.4 & 3.4 \\
            \midrule
            \textbf{TADACap-diverse} & GPT4 & \cellcolor{pale7} 30.1 & \cellcolor{pale7}  7.3 & 27.8 & \cellcolor{pale7} 17.6 \\
            TADACap-nn  & GPT4 & \cellcolor{pale1} \bf 33.0 & \cellcolor{pale1} \bf 38.9 & \cellcolor{pale1} \bf 30.6 & \cellcolor{pale1} \bf 34.8 \\
            TADACap-random & GPT4 & 29.9 & 6.3 & \cellcolor{pale7} 28.9 & \cellcolor{pale7} 17.6 \\
            TADACap-zs & GPT4 & 17.3 & 0.8 & 6.6 & 3.7 \\
            \midrule
            \textbf{TADACap-diverse} & Vicuna & \cellcolor{pale7} 25.2 & \cellcolor{pale1} 3.9 & 19.2 & \cellcolor{pale7} 11.6 \\
            TADACap-nn  & Vicuna & \cellcolor{pale1} 27.4 & \cellcolor{pale7} 3.7 & \cellcolor{pale1}  21.8 & \cellcolor{pale7}  12.7 \\
            TADACap-random & Vicuna & 24.8 & 3.3 & \cellcolor{pale7} 19.6 & 11.4 \\
            TADACap-zs & Vicuna & 17.7 & 2.3 & 4.0 & 3.2 \\
            \midrule
            \textit{SynthStocks} & & & \\
            \midrule
            Multimodal LLM & Gemini-V & 25.1 & 4.3 & 7.4 & 5.8 \\
            Multimodal LLM & mPLUG-Owl & 19.5 & 4.4 & 7.8 & 6.1 \\
            \midrule
            \textbf{TADACap-diverse} & GPT4 & \cellcolor{pale7} 45.8 & \cellcolor{pale7} 35.6 & \cellcolor{pale7} 34.2 &  \cellcolor{pale7} 34.9 \\
            TADACap-nn  & GPT4 & \cellcolor{pale1} \bf 47.5 & \cellcolor{pale1} \bf 68.5 & \cellcolor{pale1} \bf 34.4 & \cellcolor{pale1} \bf 51.4 \\
            TADACap-random & GPT4 & 42.5 & 32.7 & 29.3 & 31.0 \\
            TADACap-zs & GPT4 & 27.0 & 5.2 & 17.6 & 11.4 \\
            \midrule
            \textbf{TADACap-diverse} & Vicuna & 27.3 & 10.8 & 16.5 & 13.7 \\
            TADACap-nn  & Vicuna & \cellcolor{pale1} 32.3 & \cellcolor{pale1} 20.2 & \cellcolor{pale1} 20.3 & \cellcolor{pale1} 20.2 \\
            TADACap-random & Vicuna & \cellcolor{pale7} 31.5 & \cellcolor{pale7} 12.8 & \cellcolor{pale7} 19.6 & \cellcolor{pale7} 16.2 \\
            TADACap-zs & Vicuna & 25.8 & 6.8 & 15.5 & 11.2 \\
            \bottomrule
            \end{tabular}
        }
    }
    \caption{Benchmark results of proposed TADACap-diverse in comparison to SOTA multimodal models, ablation variants including TADACap-nn, TADACap-random, and TADACap-zs baseline. \textbf{Bold} indicates the best performance across all models for each dataset, \colorbox{pale1}{blue} highlight denotes the best performance within the same underlying large language model (LLM), and a \colorbox{pale7}{gray} highlight indicates the second best performance within the same LLM.}
    \label{tab:benchmarks}
\end{table*}

\subsection{Datasets}\label{sec:datasets}
Due to the absence of publicly available domain-aware time-series caption datasets, we introduced four datasets covering various domains for benchmarks. These datasets include two real datasets and two synthetic datasets, as detailed below.

\textbf{\textit{RealCovid}}: This dataset contains 154 time-series that represent the number of covid positive cases over time~\cite{covid19}. We designed user surveys to gather ground truth annotations on domain-aware captions for these time series, where the users are provided a few examples of domain-aware captions in this domain, such that the captioning style remains coherent in the dataset.

\textbf{\textit{RealKnee}}: This dataset contains 112 time-series that represent the knee angle of a person during locomotion~\cite{hu2018benchmark}. Similar to \textit{RealCovid}, we gathered ground truth annotations via user surveys.

\textbf{\textit{SynthPhysics}}: This dataset contains 200 time-series that represent object velocity. This dataset was created using a mix of linear \(x = p_0 + p_1 t\), and exponential \(x = q_0 \exp(q_1 t)\) processes. Both functions describe the velocity of an object. For example, if \(p_1 > 0\), the object moves with increasing velocity or constant acceleration. If \(p_1 = 0\), the object has constant velocity and zero acceleration. In the exponential case, if \(q_1 > 0\), the velocity and acceleration increase over time. These functions are used to generate series with linearly and exponentially increasing or decreasing velocity. We automatically generate the ground-truth domain-aware captions based on the values of the parameters. For example, a linear process with positive $p_1$ could result in a caption conveying "linear velocity increase". 

\textbf{\textit{SynthStock}}: This dataset contains 200 time-series that represent financial stock price series. This dataset was generated using a discrete mean-reverting process, adopted from~\cite{byrd2019explaining}: 
\begin{align*}
r_t = \max\{0, \kappa \bar{r} + \left(1 - \kappa\right) r_{t-1} + u_t\}, \quad r_0 = \bar{r},
\end{align*}
where \(\bar{r}\) is the mean value of the time series, \(\kappa\) is the mean-reversion parameter, and \(u_t \sim \mathcal{N}(0, \sigma^2)\) is random noise added to the time series at each time step \(t\). Trend \(T\) is added to \(r_t\) at each time step \(t\) to indicate the incline or decline of the stock value. Megashocks represent infrequent exogenous events with significant impacts on the generating process. Megashocks can occur at any time \(t\) with probability \(p\) and are drawn from \(\mathcal{N}(0, \sigma_{\text{shock}}^2)\) where \(\sigma_{\text{shock}} \gg \sigma\). This setup allows for various scenarios such as up/down trends, frequent/infrequent shocks, and strong/medium/low volatility. Similar to \textit{SynthPhysics}, we automatically generate the ground-truth domain-aware captions based on the values of the parameters. For instance, as the trend \textit{T} approaches zero, the caption could convey "the stock price remains stable", whereas higher values of $\sigma$ could result in a caption about the stock experiencing significant volatility.

\subsection{Benchmarks}~\label{sec:benchmarks}
We benchmarked the performance of TADACap with our proposed diverse retrieval strategy (\textbf{TADACap-diverse}), against state-of-the-art free multimodal models including \textbf{Gemini-V} and \textbf{mPLUG-Owl}, and ablation variants of TADACap, as introduced below. We used GPT-4 and Vicuna as the LLM in our framework (Figure~\ref{fig:overview}). For each dataset, we iterate through each image as the query (excluding the selected $k$ diverse samples in the case of TADACap-diverse), and use the rest of the dataset as the target domain database.

\textbf{TADACap-nn}: Retrieves the nearest-neighbor samples from the target domain database instead of diverse samples, which resembles in SmallCap~\cite{ramos2022smallcap}. As discussed in Section~\ref{sec:related_works}, SmallCap is the first work that leverages retrieval augmentation for training-free domain transfer in image captioning, achieving SOTA performance on the domain-transfer image captioning task. However, SmallCap or TADACap-nn relies on annotations for all images in the target domain database. In contrast, TADACap-diverse reduces manual efforts by requiring only annotations on a few diverse samples.

\textbf{TADACap-random}: Retrieves random samples from the target domain database, serving as an ablation variant in our benchmark.

\textbf{TADACap-zs}: A zero-shot baseline that does not leverage retrievals. Given the domain-agnostic caption, the LLM generates the domain-aware captions based on a generic description of the domain itself. For example, in the \textit{RealCovid} dataset, LLM is prompted with "Translate the time-series description 'xxx' in the context of number of Covid cases," where 'xxx' is the domain-agnostic caption. This technique relies on LLM's inherent ability to understand the described domains.

\textbf{Gemini-V} (Gemini 1.0 Pro Vision)~\cite{team2023gemini} is a foundational large language model under the Gemini series that integrates capabilities from both text and visual modalities. This model is designed to understand inputs that include text and images, enabling it to generate relevant text responses based on a comprehensive analysis of the combined data. We prompted the model with "Describe the time-series in the context of xxx", where xxx provides the context of the time-series domain. For example, in the \textit{SynthStock} case, we prompted the model with "Describe the time-series in the context of stock price series."

\textbf{mPLUG-Owl} is a large language model known for its multi-modal capabilities, capable of generating captions for images in the open world. It is based on the state-of-the-art mPLUG~\cite{li-etal-2022-mplug}, the first model to achieve human parity on the VQA Challenge~\cite{antol2015vqa}. The model is prompted in a manner similar to Gemini-V. For instance, to analyze time-series data, it is prompted with "Describe the time-series in the context of xxx", where "xxx" specifies the domain context of the time-series.

We used well-established metrics, including ROUGE~\cite{lin-2004-rouge}, CIDEr \cite{vedantam2015cider}, SPICE~\cite{anderson2016spice}, and SPIDEr~\cite{liu2017improved}. These metrics assess various aspects of the generated captions, including syntactic alignment and semantic accuracy compared with ground truth captions. 

\subsection{Results}

We benchmarked all methods against the four introduced test datasets: \textit{RealCovid}, \textit{RealKnee}, \textit{SynthStock}, and \textit{SynthPhysics}, as summarized in Table~\ref{tab:benchmarks}. We also showcase qualitative examples in Figure~\ref{fig:datasets_results}.

The results indicate that TADACap variants (TADACap-diverse, TADACap-nn, TADACap-random) outperform zero-shot multimodal LLMs (Gemini-V and mPLUG-Owl), demonstrating that adding ICL to unimodal LLMs enhances performance, especially in niche domains like the \textit{RealKnee} dataset. This finding emphasizes the importance of providing context when working with datasets from specialized domains. While our approach primarily focuses on integrating CLIP with unimodal LLMs using an effective retrieval strategy, it is worth noting that the methodology can also be extended to multimodal LLMs. This paper aims to showcase that even with unimodal LLM, leveraging effective retrieval strategies can generate quality captions.

Under the same zero-shot setting, we observed that the performances of the multimodal LLMs and TADACap-zs are largely on par due to the absence of ICL. TADACap-zs benefits from domain-agnostic captions, which describe the shape of time-series in the image as part of the TADACap pipeline. This advantage is particularly evident in datasets with significant shape changes, such as \textit{SynthStock}. As shown in \textit{SynthStock} example in Figure~\ref{fig:datasets_results}, TADACap-zs tend to outperform the multimodal LLMs with more detailed shape descriptions.

Among TADACap variations, when GPT-4 is used as the LLM in Figure~\ref{fig:overview}, TADACap-diverse performs on par with TADACap-nn in terms of SPICE, a metric for semantic accuracy. However, TADACap-diverse does not perform as well on other metrics focused on syntactic alignment with the ground truth. Similarly, when employing Vicuna as the LLM in Figure~\ref{fig:overview}, TADACap-diverse continues to demonstrate comparable performance to TADACap-nn on semantic accuracy. The ranking of the TADACap variants fluctuates more compared to GPT-4, due to Vicuna's less robust adaptation through ICL. 

For time series captioning, semantic accuracy is more crucial than syntactic alignment to accurately describe the shapes and patterns in the time series. Both GPT-4 and Vicuna, when used as the LLM in our framework, allow TADACap-diverse to achieve similar semantic accuracy (SPICE) compared to TADACap-nn. Note that TADACap-nn resembles SOTA method SmallCap~\cite{ramos2022smallcap}, as explained in Section~\ref{sec:benchmarks}. Yet, TADACap-diverse requires significantly fewer annotations. Specifically, TADACap-diverse needs $k$ annotations on the diverse samples, which are much less than the $N$ annotations required for the complete dataset in the case of TADACap-nn. Our approach reduces the annotation burden while maintaining on-par semantic accuracy, showcasing the potential of TADACap-diverse as a more efficient alternative to the nearest-neighbor approach. TADACap-diverse is effective because the retrieved samples cover diverse time-series shapes and temporal dynamics in the target domain, thus providing a comprehensive understanding of the target domain for the LLM in TADACap to generate quality captions.

\begin{figure}[!t]
    \centering
    \includegraphics[scale = 0.28]{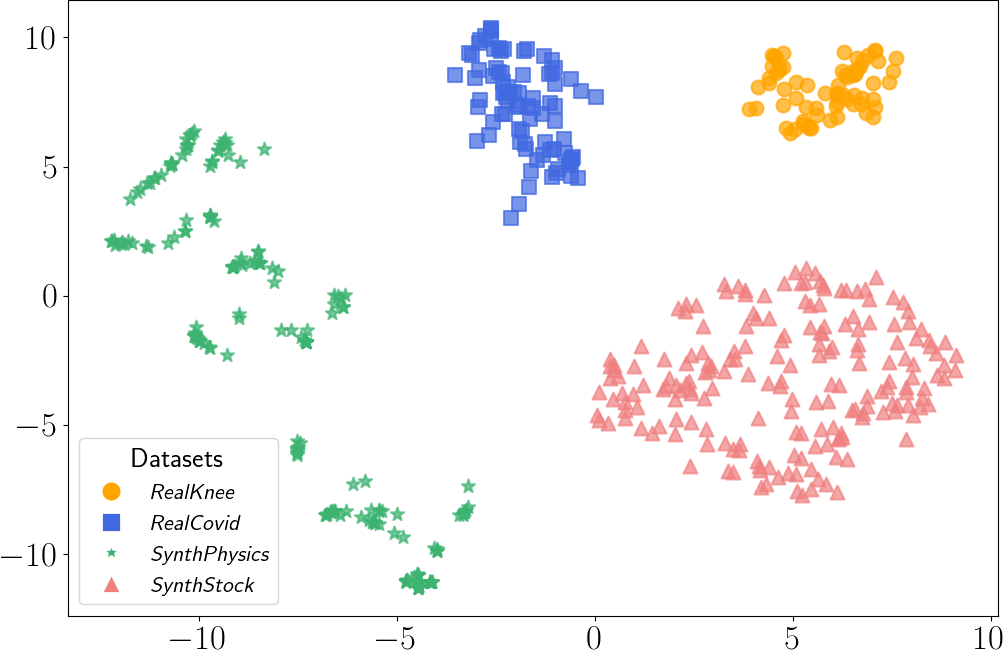}
    \caption{t-SNE visualization of image embeddings for each dataset. Colors represent the different datasets.}
    \label{fig:tsne}
\end{figure}

In addition, our four datasets encompass varying levels of diversity, as shown by the spread of the t-SNE visualization of the image embeddings of each dataset in Figure~\ref{fig:tsne}. Our findings indicate that TADACap-diverse remains robust across these different diversity levels, effectively adapting to varying domains and ensuring consistent performance.

\section{Conclusions} \label{conclusions}
In conclusion, we tackle the new challenge of domain-aware time-series captioning, addressing a gap left by previous works that focused on domain-agnostic time-series captioning. We present TADACap and, specifically, TADACap-diverse, a novel approach that retrieves diverse samples for domain-aware captioning of time-series images. Our method successfully adapts to new domains without the need for retraining by leveraging diverse image-caption pairs within a target domain database. Our experiments show that TADACap-diverse achieves comparable semantic accuracy to TADACap-nn, which resembles the SOTA method SmallCap in image captioning for novel domains. However, TADACap-diverse requires significantly less annotation effort. Notably, the robustness of TADACap-diverse across various levels of dataset diversity highlights its adaptability and efficiency. Furthermore, we have contributed new datasets to the community, facilitating further research in domain-aware time-series captioning.

\section*{Disclaimer}
This paper was prepared for informational purposes by the Artificial Intelligence Research group of JPMorgan Chase \& Co. and its affiliates (``JP Morgan'') and is not a product of the Research Department of JP Morgan. JP Morgan makes no representation and warranty whatsoever and disclaims all liability, for the completeness, accuracy or reliability of the information contained herein. This document is not intended as investment research or investment advice, or a recommendation, offer or solicitation for the purchase or sale of any security, financial instrument, financial product or service, or to be used in any way for evaluating the merits of participating in any transaction, and shall not constitute a solicitation under any jurisdiction or to any person, if such solicitation under such jurisdiction or to such person would be unlawful.

\bibliography{main}

\newpage
\appendix
\onecolumn

\section{Datasets}

\subsection{Overview of Datasets}
In Table \ref{tab:data_overview} we show an overview of the four datasets used, to real and two synthetic. In the real datasets, there is more than one caption per image. This is because we collected multiple user-generated captions for each image in the real dataset.

\begin{table*}[h!]
    \centering
\begin{tabular}{llcc}
\toprule
    & Dataset & Number of images & Number of Captions \\
\midrule
\textit{Simulated} & & & \\
\midrule
& SynthStock & 200 & 200  \\
& SynthPhysics & 200 & 200 \\
\midrule
\textit{Real} & & & \\
\midrule
    & RealCovid & 91 & 163 \\
    & RealKnee & 79 & 132 \\
\bottomrule
\end{tabular}
    \caption{Datasets information. }
    \label{tab:data_overview}
\end{table*}

\subsection{Dataset generation}
\subsubsection{SynthStock dataset}

As described in the Experiments section, we created the domain-agnostic time series dataset and the domain-aware time series caption dataset \textbf{SynthStock} using a discrete mean-reverting process, described but the equation below: 
\begin{align*}
r_t = \max\{0, \kappa\bar{r}+ \left( 1 - \kappa \right) r_{t-1} + u_t\}, r_0 = \bar{r},
\end{align*}
where $\bar{r}$ is a mean value of the time series, $\kappa$ is a mean-reversion parameter and $u_t \sim \mathcal{N} (0, \sigma^2)$ is random noise added to the time series at each time step $t$. 
We bring directionality and a possibility of a large shock occurrence to the above generating process by introducing the concepts of trend and megashocks. Trend $T$ is added to $r_t$ at each time step $t$ to indicate the incline or decline of the stock value. As in \cite{byrd2019explaining}, megashocks are intended to represent the exogenous events that occur infrequently and can have significant impact on the generating process. Mathematically, megashocks can arrive at any time $t$ with probability of occurrence $p$, and are drawn from $\mathcal{N} (0, \sigma_{shock}^2)$ where $\sigma_{shock}>>\sigma$.

To auto-generate the caption for the generated time series, we associate the numerical value of each parameter $\bar{r}$, $\kappa$, $\sigma$, $T$, $p$, $\sigma_{shock}$ with a sentiment that describes it. Below we list the different sentences used for agnostic and domain-aware captions, with their corresponding sets of parameters:

\noindent{\bf Trend up}\\
\begin{description}
    \item[\dorange{Agnostic:}] \texttt{["it goes upward", "it grows", "has a positive growth", "is rising", "is climbing"]}
     \item[\dogreen{Stock domain:}] \texttt{["the price grows", "the price increases", "the asset has a positive growth", "the price of the equity is rising", "the price is climbing"]}
    \item[\doblue{Parameters: }]\texttt{"mean":[90,95] ,"sigma":[0.001,0.02], "p":[0.0005, 0.0006], "T":[0.5, 0.6], "kappa":[0.001, 0.005], "shock sigma":[0.001, 0.005]}
\end{description}

\noindent{\bf Trend neutral}\\
\begin{description}
    \item[\dorange{Agnostic:}] \texttt{["is neutral", "is horizontal", "is non-increasing", "is flat", "is stable", "the slope is unchanged"]}
     \item[\dogreen{Stock domain:}] \texttt{["the stock price remains neutral", "the price is flat", "the stock value is stable", "the price is unchanged"]}
    \item[\doblue{Parameters: }] \texttt{"mean":[95,105], "sigma":[0.001,0.02], "p":[0.0005, 0.0006], "T":[-0.01,0.01], "kappa":[0.001, 0.005], "shock sigma":[0.001, 0.005]}
\end{description}
    
\noindent{\bf Trend down}\\
\begin{description}
    \item[\dorange{Agnostic:}] \texttt{["the stock is declining", "the price is falling", "the price is sliding", "the stock is plummeting", "the equity has a downward slope"]}
    \item[\dogreen{Stock domain:}] \texttt{["the stock is declining", "the price is falling", "the price is sliding", "the stock is plummeting", "the equity has a downward slope"]}
    \item[\doblue{Parameters: }] \texttt{"mean":[105,110], "sigma":[0.001,0.02], "p":[0.0005, 0.0006], "T":[-0.6,-0.5], "kappa":[0.005, 0.010], "shock sigma":[0.001, 0.005]}
\end{description}

\noindent{\bf Shock high}\\
\begin{description}
    \item[\dorange{Agnostic:}] \texttt{["has frequent shocks", "shows common jumps"]}
    \item[\dogreen{Stock domain:}] \texttt{["the stock experiences frequent shocks", "the price shows jumps"]}
    \item[\doblue{Parameters: }] \texttt{"mean":[96,104], "sigma":[0.005, 0.010], "p":[0.1, 0.3], "T":[-0.02,0.02], "kappa":[0.01, 0.050], "shock sigma":[0.03, 0.07]}
\end{description}

\noindent{\bf Shock low}\\
\begin{description}
    \item[\dorange{Agnostic:}] \texttt{["has infrequent jumps", "shocks are uncommon", "jumps are rare"]}
    \item[\dogreen{Stock domain:}] \texttt{["the price does not experience high fluctuations", "price jumps are uncommon", "stock value jumps are rare"]}
    \item[\doblue{Parameters: }] \texttt{"mean":[90,110], "sigma":[0.005, 0.010], "p":[0.0005, 0.05], "T":[-0.02,0.02], "kappa":[0.01, 0.050], "shock sigma":[0.001, 0.005]}
\end{description}

\noindent{\bf Sigma high}\\
\begin{description}
    \item[\dorange{Agnostic:}] \texttt{["has strong variability", "has significant variations", "has aggressive variations", "is unstable", "has high fluctuation", "is noisy", "is variable"]}
    \item[\dogreen{Stock domain:}] \texttt{["the price shows high volatility", "has significant volatility", "has aggressive variations in price", "the value is unstable", "price shows has high fluctuation", "is volatile"]}
    \item[\doblue{Parameters: }] \texttt{"mean":[98,102], "sigma":[0.02, 0.035], "p":[0.0005, 0.05], "T":[-0.01,0.01], "kappa":[0.01, 0.05], "shock sigma":[0.001, 0.005]}
\end{description}

 \noindent{\bf Sigma medium}\\
\begin{description}
    \item[\dorange{Agnostic:}] \texttt{["has moderate variability", "has intermediate variance", "has medium variability"]}
    \item[\dogreen{Stock domain:}] \texttt{["the stock experiences moderate volatility", "has intermediate volatility", "prices have medium variability"]}
    \item[\doblue{Parameters: }] \texttt{"mean":[95,105], "sigma":[0.005, 0.020], "p":[0.0005, 0.05], "T":[-0.01, 0.01], "kappa":[0.005, 0.010], "shock sigma":[0.001, 0.005]}
\end{description}
  
\noindent{\bf Sigma low}\\
\begin{description}
    \item[\dorange{Agnostic:}] \texttt{["has small variability", "shows a slight variability", "has negligible variations", "has weak variations", "is stable", "is smooth"]}
    \item[\dogreen{Stock domain:}] \texttt{["has small volatility", "the stock shows a slight variability", "the stock has negligible volatility", "has low volatility", "the price remains stable"]}
    \item[\doblue{Parameters: }] \texttt{"mean":[90,110], "sigma":[0.001, 0.005], "p":[0.0005, 0.05], "T":[-0.01,0.01], "kappa":[0.001, 0.005], "shock sigma":[0.001, 0.005]}
\end{description}

\subsubsection{SynthPhysics dataset}

Following a similar procedure as before, we generated the SynthPhysics dataset using two types of generating functions, linear: $x = p_0 + p_1 t$ and exponential: $x=q_0 \exp(q_1 t)$. Both functions are used to describe the velocity of an object, where for example if $p_1 > 0$ we can say that the object moves with increasing velocity, or equivalently, with constant acceleration, or in the case of $p_1=0$, the object is described as having constant velocity and null acceleration. In the exponential case, for example if $q_1>0$ and $q_1>0$ we can say that the velocity increases, or that the acceleration increases over time. 
To auto-generate the caption, we associate phrases to the sign of the parameters. 
Below we list some examples of sentences used:\\
\noindent{\bf Exponential Positive}\\
\begin{description}
    \item[\doblue{Parameters: }] \texttt{$q_1>0$, $q_1>0$}
   \item[\dogreen{Velocity:}] \texttt{["The velocity/The speed grows/increases exponentially"],["The speed/velocity/acceleration of the object increases over time"], ["The object moves with exponentially increasing velocity and acceleration"]}
\end{description}

\noindent{\bf Exponential Negative}\\
\begin{description}
    \item[\doblue{Parameters: }] \texttt{$q_1>0$, $q_1<0$}
    \item[\dogreen{Velocity:}] \texttt{["The velocity/The speed decreases exponentially"],["The speed/velocity/acceleration of the object decreases over time"], ["The object moves with exponentially decreasing velocity and acceleration"]}
\end{description}

In order to generate more complex descriptions, we randomly sample two sets of parameters and combine the functions, creating a two-sentence caption. For example the resulting caption can be: "The object moves with exponentially decreasing velocity in the first part. It has constant velocity afterwards", or "The speed increases uniformly over time at the beginning. It decreases exponentially in the end".

\section{Additional Results}

\subsection{Diverse $k$ Samples Selection}
\begin{figure*}[h!]
\centering
\subfigure[]{
  \includegraphics[width=0.35\textwidth]{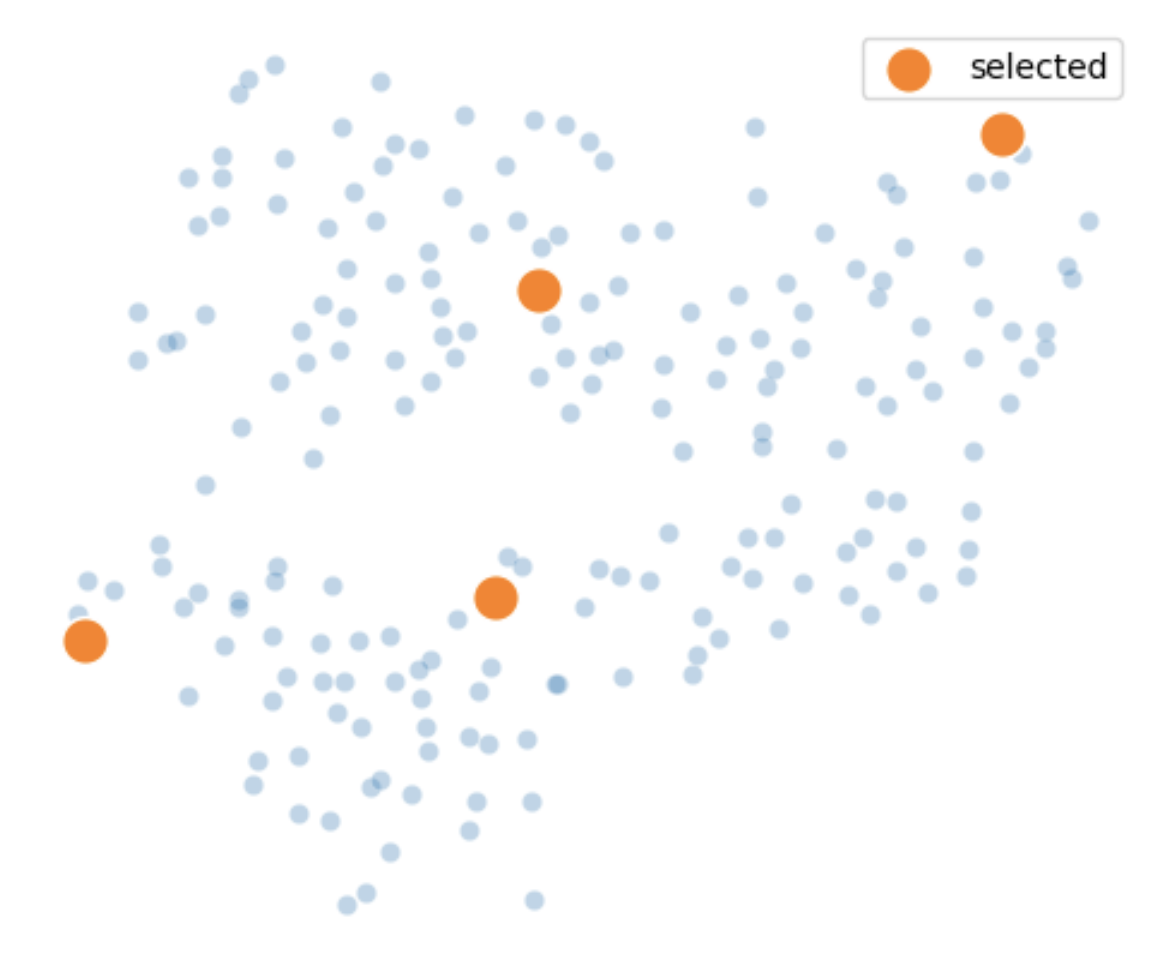}
}
\subfigure[]{
  \centering
  \includegraphics[width=0.47\textwidth]{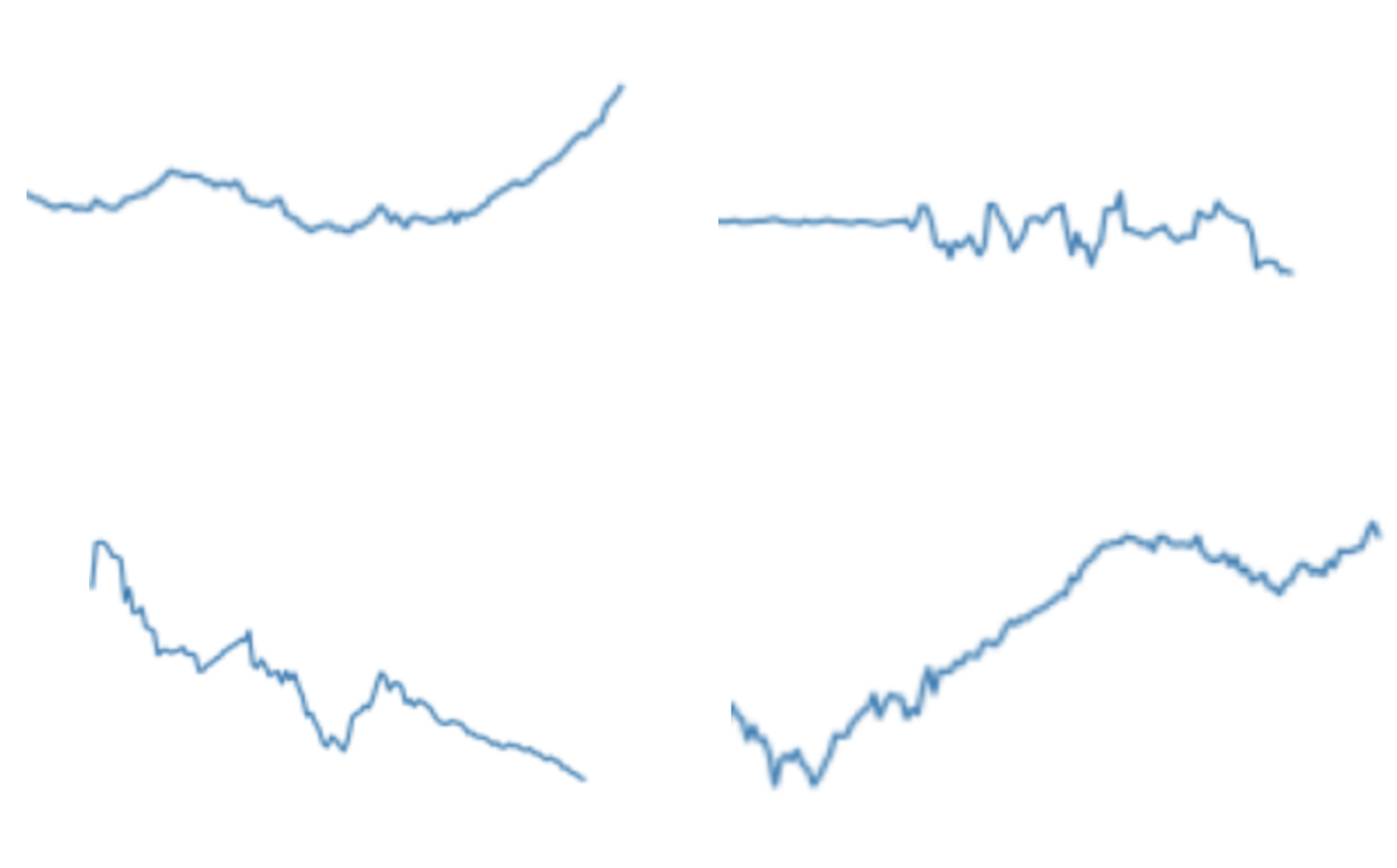}
}
    
\caption{Example of selected diverse $k=4$ samples in our test database \textbf{SynthStock}. (a) TSNE visualization of data distribution of CLIP image embeddings of images in the database, \textit{orange} marks the selected samples; (b) selected diverse images.}
\label{fig:diverse_k_stock}
\end{figure*}

\begin{figure*}[h!]
\centering
\subfigure[]{
  \includegraphics[width=0.4\textwidth]{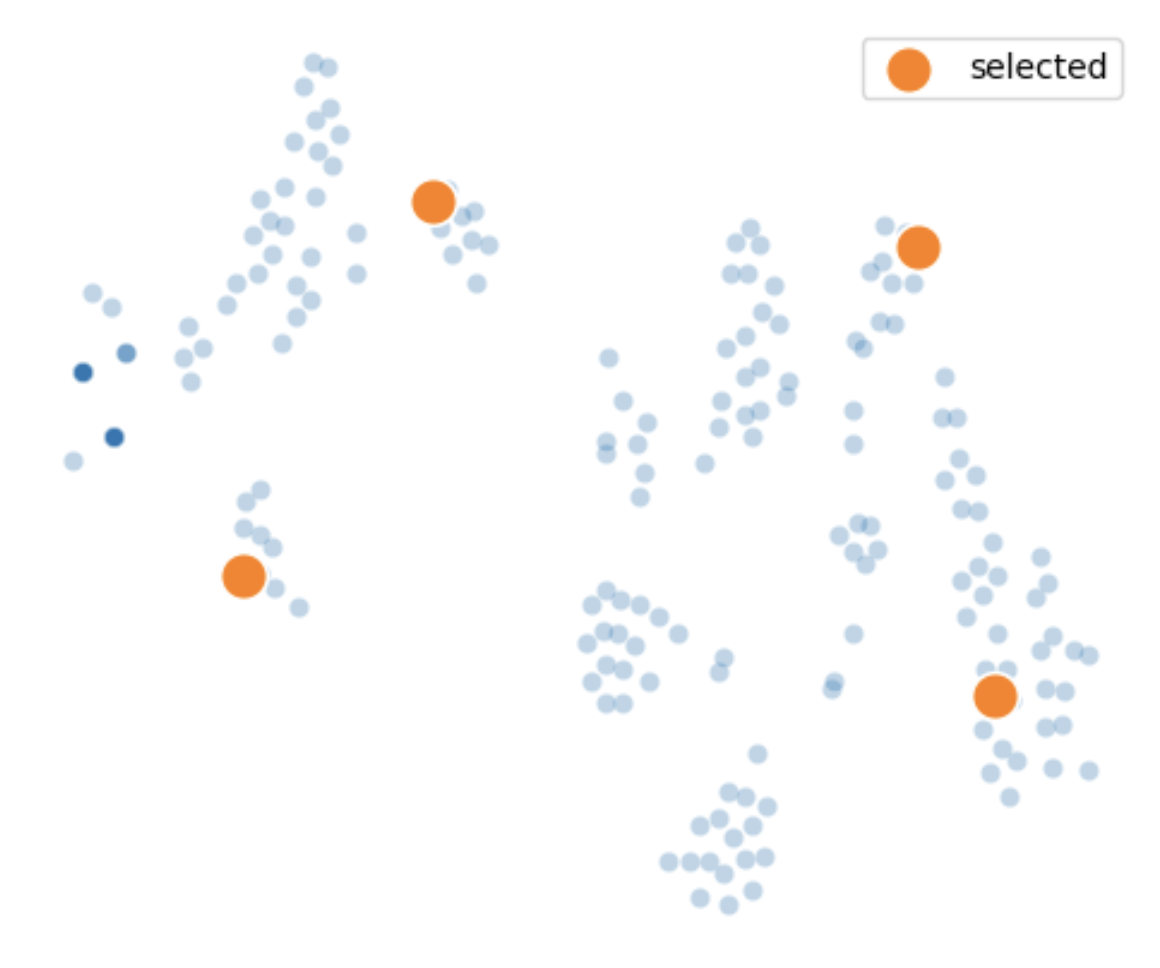}
}
\subfigure[]{
  \centering
  \includegraphics[width=0.47\textwidth]{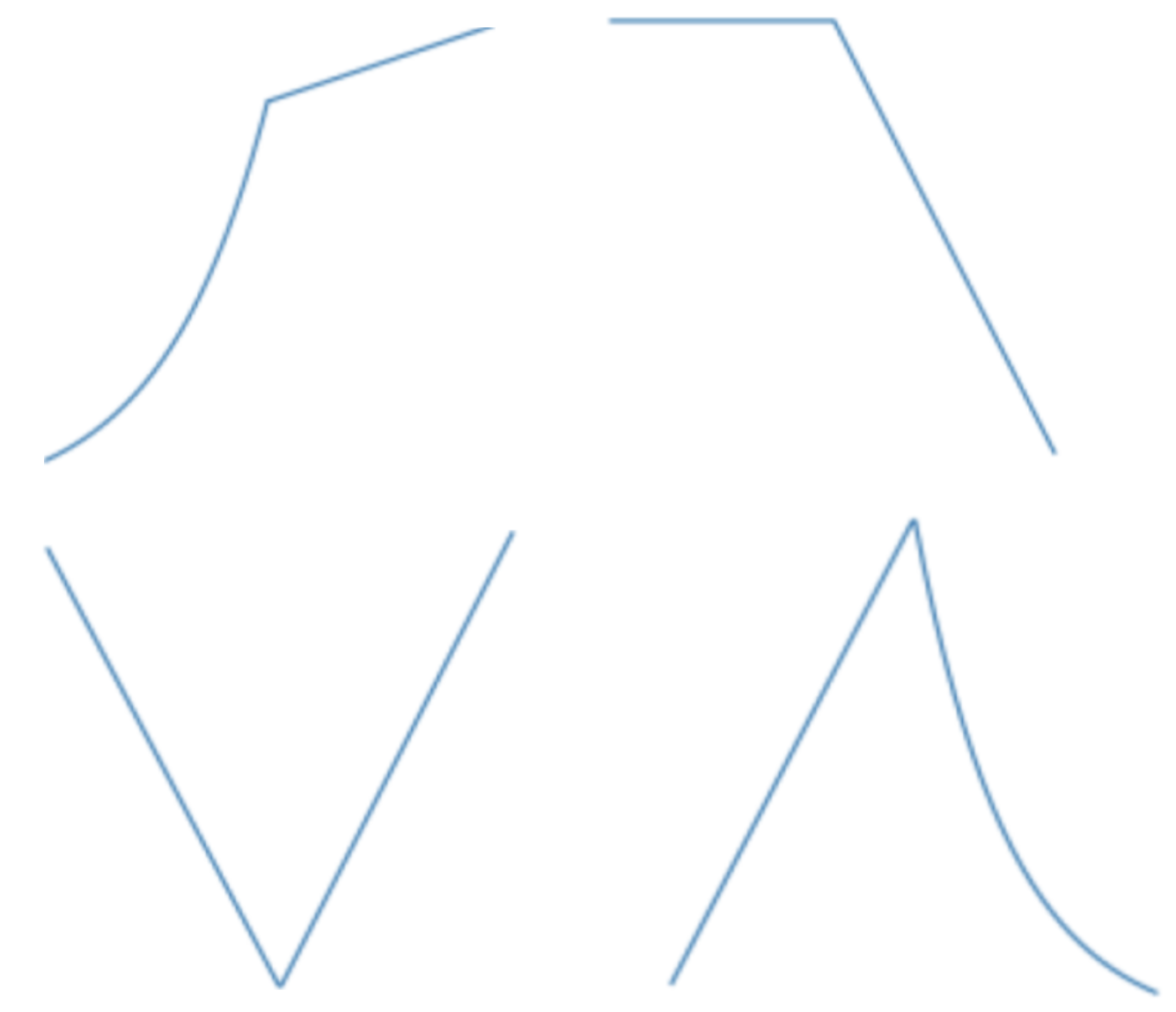}
}
    
\caption{Example of selected diverse $k=4$ samples in our test database \textbf{SynthPhysics}. (a) TSNE visualization of data distribution of CLIP image embeddings of images in the database, \textit{orange} marks the selected samples; (b) selected diverse images.}
\label{fig:diverse_k_physics}
\end{figure*}

\begin{figure*}[h!]
\centering
\subfigure[]{
  \includegraphics[width=0.35\textwidth]{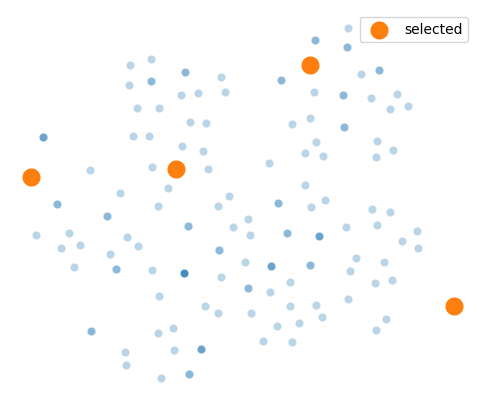}
}
\subfigure[]{
  \centering
  \includegraphics[width=0.45\textwidth]{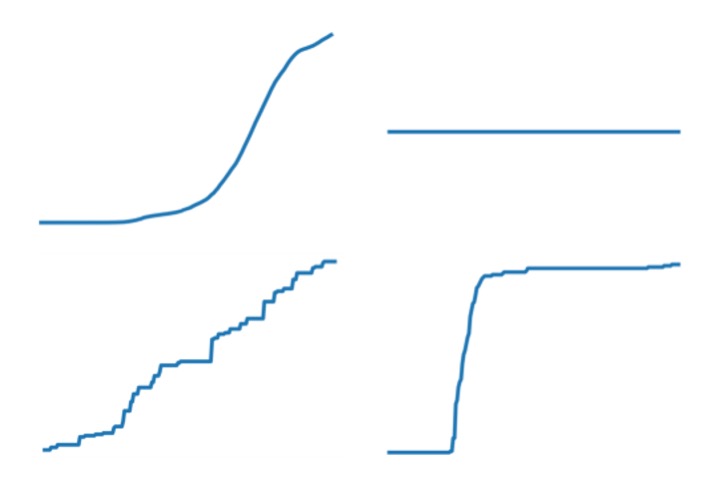}
}
    
\caption{Example of selected diverse $k=4$ samples in our test database \textbf{RealCovid}. (a) TSNE visualization of data distribution of CLIP image embeddings of images in the database, \textit{orange} marks the selected samples; (b) selected diverse images.}
\label{fig:diverse_k_covid}
\end{figure*}

\begin{figure*}[h!]
\centering
\subfigure[]{
  \includegraphics[width=0.35\textwidth]{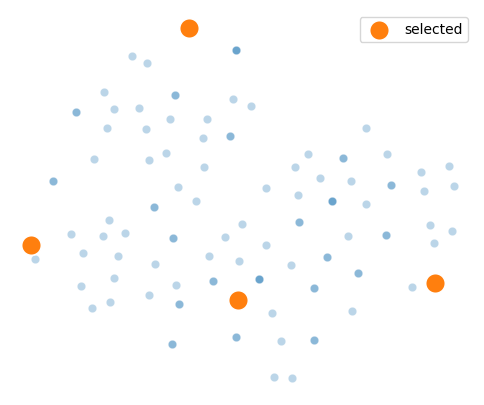}
}
\subfigure[]{
  \centering
  \includegraphics[width=0.45\textwidth]{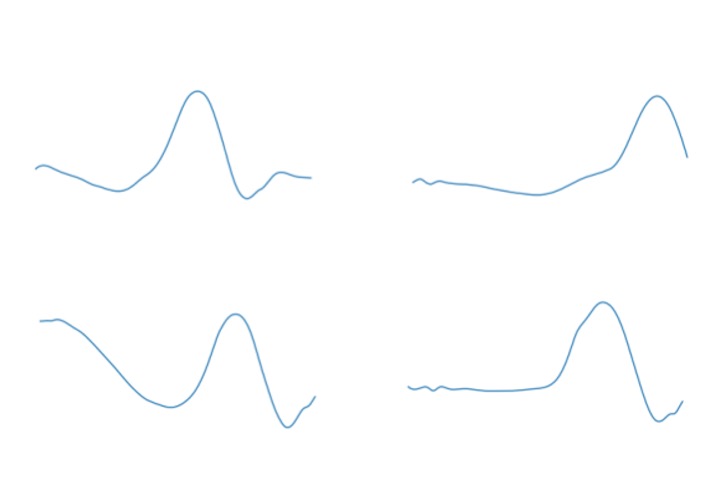}
}
    
\caption{Example of selected diverse $k=4$ samples in our test database \textbf{RealKnee}. (a) TSNE visualization of data distribution of CLIP image embeddings of images in the database, \textit{orange} marks the selected samples; (b) selected diverse images.}
\label{fig:diverse_k_knee}
\end{figure*}

Figures~\ref{fig:diverse_k_stock} and~\ref{fig:diverse_k_physics} depict the TSNE visualizations of CLIP image embeddings from the \textbf{SynthStock} and \textbf{SynthPhysics} datasets, emphasizing the selection of diverse samples with $k=4$ (marked in orange) within the distribution. These visualizations are complemented by their respective corresponding images. Similarly, Figure~\ref{fig:diverse_k_covid} provides a comparable visualization for the \textbf{RealCovid} dataset. Moving forward to Figure~\ref{fig:diverse_k_knee}, it showcases the TSNE visualization of CLIP image embeddings from the \textbf{RealKnee} dataset, featuring the selection of diverse $k=4$ samples (highlighted in orange), in conjunction with their corresponding images. These visualizations unequivocally demonstrate that the chosen diverse samples exhibit both visual diversity, as revealed by the TSNE plots, and a diversity of shapes within their respective images.

\end{document}